\PassOptionsToPackage{hyphens}{url}
\documentclass[reprint,
prl,
amsmath,amssymb,
aps,
]{revtex4-2}

\usepackage{graphicx}
\usepackage{dcolumn}
\usepackage{bm}

\usepackage{amsmath}
\usepackage{amssymb}
\usepackage{amsthm}
\usepackage{physics}
\usepackage{hyperref}
\usepackage{xcolor}
\definecolor{darkgreen}{rgb}{0,0.5,0} 

\allowdisplaybreaks

\begin{document}

\preprint{}

\title{Multispin Physics of AI Tipping Points and Hallucinations}

\author{Neil F. Johnson}%
 \email{neiljohnson@gwu.edu}
\author{Frank Yingjie Huo}
\affiliation{%
 Physics Department, George Washington University, Washington, DC 20052, U.S.A.
}%

\date{\today}

\begin{abstract}
    Output from generative AI such as ChatGPT, can be repetitive and  biased. But more worrying is that this output can mysteriously tip mid-response from good (correct) to bad (misleading or wrong) without the user noticing. In 2024 alone, this reportedly caused $\$67$ billion in losses and several deaths. 
    Establishing a mathematical mapping to a multispin thermal system, we reveal a hidden tipping instability at the scale of the AI's `atom' (basic Attention head). We derive a simple but essentially exact formula for this tipping point which shows directly the impact of a user's prompt choice and the AI's training bias. We then show how the output tipping can get amplified by the AI's multilayer architecture. As well as helping improve AI transparency, explainability and performance, our results open a path to quantifying users' AI risk and legal liabilities.

\end{abstract}

\maketitle


We may not notice when the output from our generative AI (e.g. ChatGPT, GPT-5) tips mid-response from good output (correct) to bad output (plausible but misleading or wrong, i.e. hallucination). Recent examples include the apparent suicide of a 14-year-old after his trusted AI companion tipped mid-response from responsible to pro-suicide narratives \cite{reuters_setzer,ap_setzer}; a court case in which attorneys' LLM-generated briefs started off accurate but then tipped to cite fabricated legal precedents \cite{reuters_legal}; Air Canada chatbots tipping mid-conversation to  offer callers bereavement refunds \cite{cbs_aircanada,guardian_aircanada}; and reports of $\$67$ billion in  damages during 2024 alone \cite{billions_losses}. 
Given that medical entities, businesses, law firms, governments and militaries are now starting to fine-tune their own agentic AI -- and given that the next generation often trusts AI's advice over that of humans \cite{teens} -- harms and lawsuits from unnoticed good-to-bad output tipping look set to skyrocket globally across medical, mental health, financial, commercial, government and military AI domains. 

There will surely never be a mathematical theory that can account for all the complexities  of ChatGPT, Claude, Gemini etc. and hence fully explain their output. On the other hand, despite myriad design differences, these distinct `Transformer' machines \cite{attention_review} all show occasional good-to-bad output tipping. This suggests that a much-needed science of output tipping does not have to account for all the multilayer architecture details. This motivates us to instead start with the `atomic' building block from which they are all built: the Attention head \cite{vaswani2023attentionneed}. 

This paper shows how and when output tipping arises at the scale of the fundamental `atom'  of any current or future Transformer-based generative AI (e.g. ChatGPT, Claude, Gemini): a basic Attention head. Establishing a mathematical equivalence to a multispin thermal problem (Fig. 1), we derive a simple formula that reveals, explains and predicts its output tipping, as well as the impact of the user's prompt choice and  training bias (Figs. 2, 3). 
Approximating the multilayer processing, we then show how the underlying multispin shifts can get amplified (Fig. 4). Mathematical details, terminology and code are in the Supporting Material (SM).

The literature already has some fascinating analyses of spin models inspired by Attention \cite{Rende23,Rende24Potts,ParisMath}, empirical analysis of AI output attractors \cite{attractors}, and AI's internal  mesoscale circuitry \cite{Nanda1, Nanda2, Nanda3, anthropic2025tracing, anthropic2025MIT, circuit_tracing_2025, lindsey2025biology}. 
This paper is separate from all these since we establish a bottom-up, multispin analysis of the most basic `atom' in generative AI like ChatGPT, and then we use this to derive a specific formula for its latent tipping instabilities.

\begin{figure}[ht]
    \centering
    \includegraphics[width=1.0
    \linewidth]{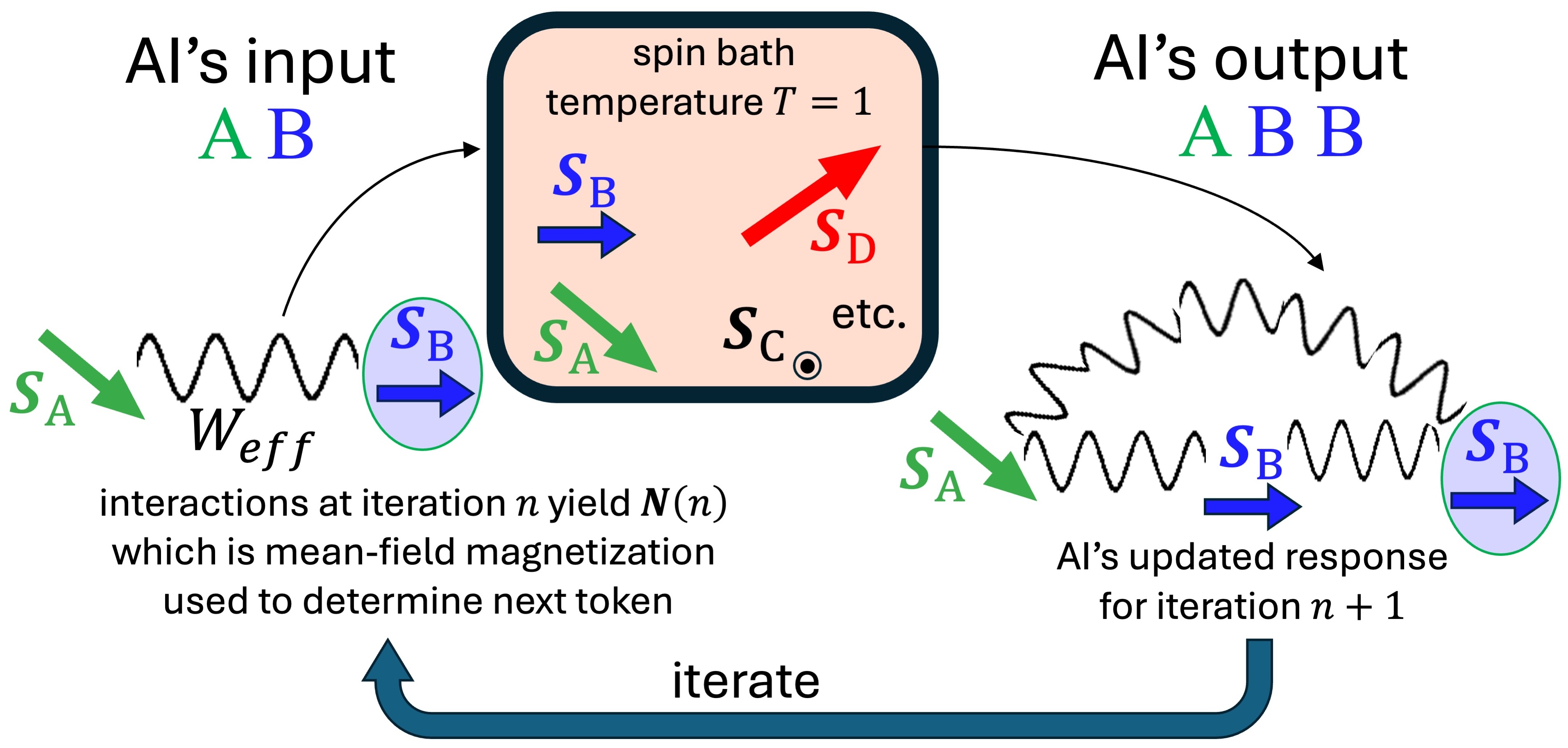}
    \caption{(a) Iterative next-token generation of generative AI such as ChatGPT (Attention head). An Attention head is mathematically equivalent to a multispin thermal system. Each spin ${\vb*{S}}_i$ represents a token (e.g. word, phrase) in an embedding space shaped by the training.}
    \label{fig:1}
\end{figure}

For all generative AI such as ChatGPT, the input (Fig. 1) gets converted into a string of tokens $\mathtt{A,B,\dots}$ etc. Each token is a spin ${\vb*{S}}_A, {\vb*{S}}_B,\dots$ in a $d$-dimensional embedding space shaped by the training phase. This input string's vectors are then `transformed' by the Attention head(s) mathematics, so that they point in directions that better capture the context of the prompt and its relation to the training data. Based on these values, the next spin is chosen and the process iterated to produce a body of output (Fig. 1).
Though for simplicity we will refer to the symbols $\mathtt{A,B,..}$ (e.g. Figs. 1-4) as individual tokens (words) where $\mathtt{B}$ is `good' content and $\mathtt{D}$ is `bad', our mathematical analysis and formula are more general: each symbol could represent a cluster of similar words or phrase(s) in a coarse-grained semantic embedding space (see Fig. 2(c)). They could also cross a spectrum between good and bad, and beyond. Figure 3(a) shows successive tippings can then occur but our formula still works, i.e. it can describe each tipping with successive pairs playing the roles of $\mathtt{B}$ and $\mathtt{D}$. Our results also hold if $\mathtt{B}$ and $\mathtt{D}$ represent two camps of thought, such as `non-woke' and `woke' (e.g. DEI) content using the language of the recent U.S. Presidential AI Executive Order 
\cite{govtai}.

The following Attention mathematics forms the key part of all generative AI such as ChatGPT \cite{vaswani2023attentionneed}. There is no fundamental theory for why it works so well, hence its choices can appear somewhat bewildering -- however we can provide an exact physics interpretation. 
First it calculates the interaction (i.e. dot-product) between the last input spin (`query') $f$ and every input spin $i$ (`key'). Operating on each with fixed training stage matrices $W_{q,k}$  can improve performance, but our spins can be seen as the result of this operation. (SM Fig. 1 confirms tipping points still occur even if we allow for different $W_{q,k}$). Each interaction is the negative of a 2-body Hamiltonian
$H(\vb*{S}_f,\vb*{S}_i) = - \vb*{S}_f \cdot \vb*{S}_i$. It can be scaled by a constant $\sqrt{d}$ without changing our conclusions. Then a thermal average is taken at fixed temperature $T=1$ (so-called Softmax) yielding 
${a_{fi}=e^{-H(\vb*{S}_f,\vb*{S}_i)}}/{\sum_{j=1}^f e^{-H(\vb*{S}_f,\vb*{S}_j)}}$. 
The mean-field magnetization $\vb*{N}{(n)}=\sum_i^f a_{fi} \vb*{S}_i$ is then calculated over the input spins (`value'), hence it embodies the last spin's `context' with respect to the input string and the training. The next token (e.g. $\mathtt{B}$ in Fig. 1) is then chosen according to the size of $\vb*{N}{(n)}$'s interaction with each possible spin in the system ($\vb*{S}_{A,B,C,D,\dots}$) and hence the ordering of their effective energy levels (Fig. 2(a)). The lower the energy level, the higher the probability that token is the next token. To allow users to choose the output's degree of surprise (stochasticity) a temperature dial $T'$ is often added which is equivalent to placing this multilevel spin system in a heat bath. Since $T'>0$ does not affect the Attention mathematics, we set $T'$ to be smaller than the level spacings. Hence the next output token is the one with the lowest energy level, i.e. `greedy decoding'. Different Attention heads can pay attention to different spin-vector components and hence different aspects of the input, e.g. adjectives versus nouns. Positional encoding can be added (e.g. periodicity) but studies show this is not strictly necessary since the  self-Attention described above can play this role \cite{haviv2022positional}.

\begin{figure}[ht]
    \centering
    \includegraphics[width=1.0\linewidth]{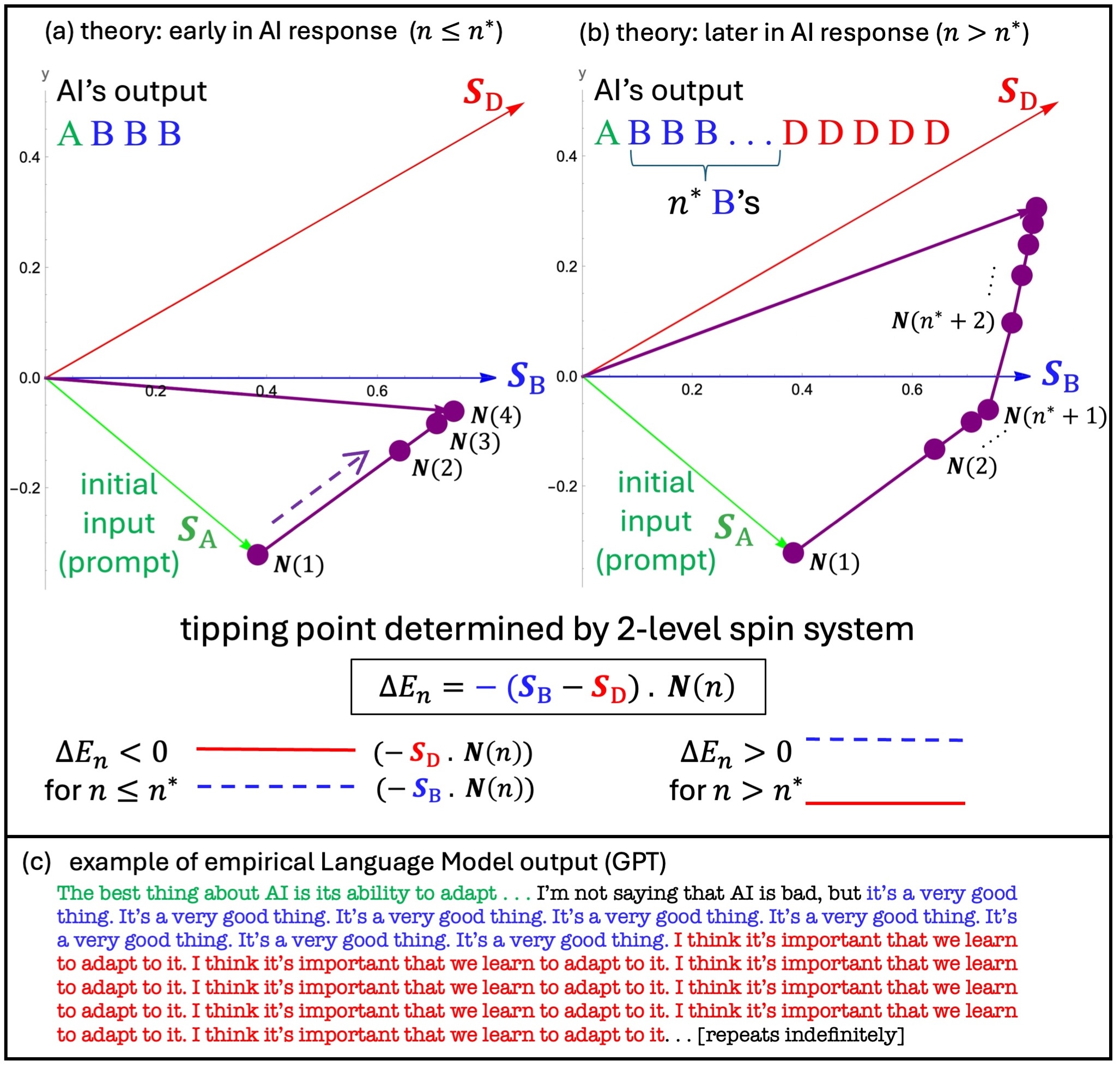}
    \caption{(a) and (b): Output tipping point at iteration $(n^*+1)$ midway through a response to a prompt, due to a transition in the identity of the largest multispin interaction. The predicted $n^*$ from our derived formula (Eq. 1) is always identical to the empirical value for a basic Attention head (see SM code). Here ${\vb*{S}}_A = (0.383,-0.321,0),\ {\vb*{S}}_B=(0.820,0,0),\ {\vb*{S}}_C = (0,0,0.500),\ {\vb*{S}}_D=(0.866,0.500,0)$. For a user's prompt $\mathtt{A}$, Eq. 1 yields $n^*=3$. The spin ${\vb*{N}}$ is a mean-field magnetization of the input spins. (c) Empirical output tipping in a full LLM (GPT-2, low $T'$),  from one type of phrase (playing the role of   $\mathtt{B}$) to another type of phrase (playing the role of $\mathtt{D}$) as in (a). SM has all derivations, code and shows the tipping's robustness to LLM-specific variations, e.g. non-identity $\mathsf{W}_{q,k}$ training phase matrices and $T'>0$. 
    }
    \label{fig:2}
\end{figure}

Figure 2 illustrates all this for a small-$d$ case where the user's prompt is simple, benign (i.e. type $\mathtt{A}$ content) and short, i.e. the input is just $\vb*{S}_{A}$. The prompt does not contain any $\mathtt{B}$  (`good' content, e.g. factually rich) or  $\mathtt{D}$ (`bad' content, e.g. wrong). 
For iteration $n=1$, the mean magnetization $\vb*{N}(1)=\vb*{S}_{A}$. Since $\vb*{N}{(1)}$'s interaction with $\vb*{S}_{B}$ is greater than with $\vb*{S}_{A,C,D}$ (i.e. $\vb*{S}_B \cdot \vb*{N}(1) >\vb*{S}_{A,C,D} \cdot \vb*{N}(1)$) the next token generated is $\mathtt{B}$. The new input for iteration $n=2$ becomes $\mathtt{A B}$, hence the mean magnetization $\vb*{N}(2)$ now averages over $\vb*{S}_{A}$ and $\vb*{S}_{B}$. So $\vb*{N}(2)$ shifts from $\vb*{N}(1)$ towards $\vb*{S}_{B}$ as shown. This process then keeps repeating.  

One might think that the generation of type $\mathtt{B}$ output would continue indefinitely in this Fig. 2 example, i.e. $\mathtt{A B B B B \dots}$. But that is not what happens. Remarkably, it suddenly shifts to $\mathtt{D}$ even though $\vb*{N}(n)$ is getting progressively closer to $\vb*{S}_{B}$ (Fig. 2(a)). This is because  there is a critical iteration number  $n=n^*$ at which $\vb*{N}(n)$ now has the largest interaction with $\vb*{S}_D$, i.e. $\vb*{S}_D \cdot \vb*{N}(n) > \vb*{S}_B \cdot \vb*{N}(n)$, meaning the lowest effective energy level becomes $H=-\vb*{S}_D\cdot \vb*{N}(n)$. Hence the generated output tips to $\mathtt{D}$. A user's choice of finite $T'$ simply broadens this transition. 

The important practical implication of this transition is that there is a sudden tipping to misleading, wrong, offensive,  dangerous or illegal content (i.e. type $\mathtt{D}$) within a single AI response that was, until then, completely good (all type $\mathtt{B}$) and which was generated by a prompt that was  benign (type $\mathtt{A}$). None of the existing AI guardrails or safety tools would have kicked in prior to this first bad output (i.e. $\mathtt{D}$) appearing. Figure 2(c) shows an empirical example of this switching in GPT-2, from a phrase $\mathtt{B}$ being repeated to a phrase $\mathtt{D}$ being repeated (N.B. we avoid giving an unpleasant example).

By calculating when $\Delta E$ changes sign (Fig. 2(a)), we can derive the following exact formula for output tipping points for any prompt size and composition, any size of vocabulary, and any size of embedding dimensions. (See SM for step-by-step algebra).  
The tipping point to $\mathtt{D}$ type output given an initial prompt $\mathtt{P_1 P_2}$ etc., will occur immediately after this number of $\mathtt{B}$ outputs:
\begin{equation}
n^{*} =
\frac{
\sum^{{{\vb*S_{P}\in \text{prompt}}}} \Bigl({{\vb*S_{P}}} \cdot {{\vb*S_{B}}} - {{\vb*S_{P}}}\cdot {{\vb*S_{D}}}\Bigr) \exp\bigl({{\vb*S_{P}}} \cdot {{\vb*S_{B}}} \bigr)
}{
\Bigl({{\vb*S_{B}}} \cdot {{\vb*S_{D}}} - {{\vb*S_{B}}}\cdot {{\vb*S_{B}
}}\Bigr) \exp\bigl({{\vb*S_{B}}} \cdot {{\vb*S_{B}}}\bigr)
}
\end{equation}
where the right-hand side is rounded to the next highest integer to produce $n^{*}$. The output tipping point $n^{*}$ is hence `hard-wired' from the moment it starts iterating a response because all its vectors and dot-products in Eq. 1 are determined by the AI's prior training and the user's prompt. 
For a user prompt $\mathtt{P}=\mathtt{A}$, Eq. 1 yields $n^*=3$ so the output is $\mathtt{A B B B D D D \dots }$ as seen empirically in Fig. 2(a). For a prompt $\mathtt{P_1 P_2 P_3 P_4}=\mathtt{ACCA}$, Eq. 1 yields $n^*=6$.  
If the prompt string is replaced by a single net spin ${\vb*S_{P}}$, Eq. 1 is well approximated by 
$\exp(({\vb*S_{P}}-{\vb*S_{B}})\cdot {\vb*S_{B}})\ [{\vb*S_{P}} \cdot ({{\vb*S_{B}}} - {\vb*S_{D}})]
/
 [{\vb*S_{B}} \cdot ({\vb*S_{D}} - {\vb*S_{B}})]
$.

Equation 1 is general in that (1) it applies to any number of embedding dimensions $d$ since changing $d$ simply changes the number of vector components; hence it can be used for any current or future ChatGPT-like AI.  
(2) It accounts for the generative AI's training (and hence its training data) via the embedding vector components for $\mathtt{A,B,C,D\dots}$; hence it can be used to explore the impact of training bias on the output. (3) It applies to any prompt by the user; hence it can be used to evaluate the impact of, for example, verbose vs. terse prompts and the effect of packing prompts with different types of content. 
Figure 3(a) shows an example of packing a prompt with content type $\mathtt{C}$ (e.g. politeness) that lies in the 2D plane. This introduces 2 consecutive tipping points: Eq. 1 describes each tipping point with the relevant energy level pairs being $\mathtt{A,B}$ then $\mathtt{B,D}$. 
(4) It applies to any size of vocabulary, since each tipping point results from a single spin pair (playing the role of $\mathtt{B}$ and $\mathtt{D}$) whose energy gap $\Delta E$ changes sign. 
As the size of the vocabulary increases for fixed $d$, there will be an increasingly complex set of tipping points as confirmed in Fig. 3(a), each described by Eq. 1. Since ChatGPT-like generative AI has a high ratio of token spin vectors to embedding dimensions (i.e. the space is crowded), the turning point topology will be very rich. 

\begin{figure}
    \centering
\includegraphics[width=1.0\linewidth]{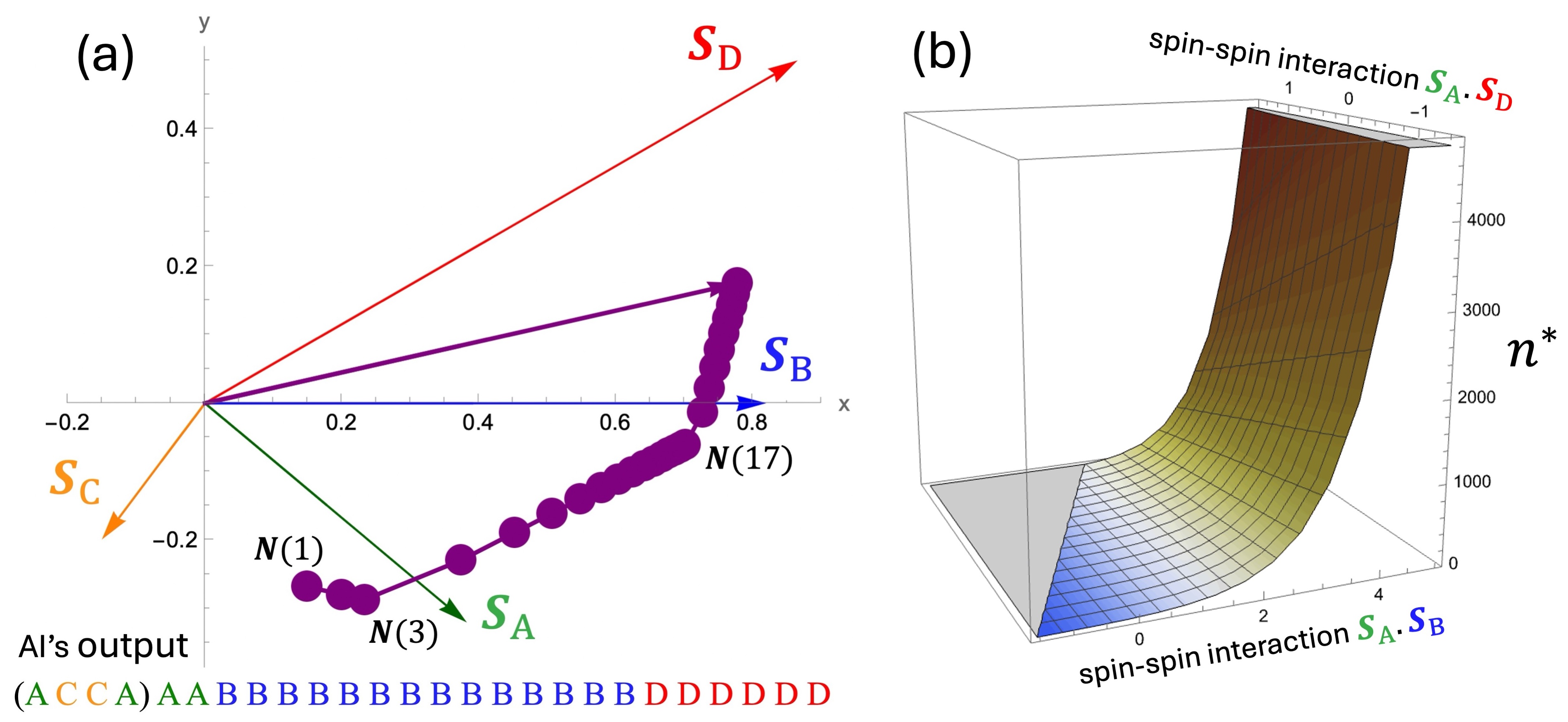}
    \caption{(a) Similar to Fig. 2(a)(b), but the more complex prompt $\mathtt{ACCA}$ with  $\mathtt{C}=(-0.150,-0.200,0)$ produces multiple tipping points. (b) Equation 1 predicts how $n^*$ (and hence the tipping point's onset) can be increased substantially by engineering the interactions between the spins for the prompt (i.e. which is  just ${\vb*S_{A}}$ for the simple prompt $\mathtt{A}$ in this example) and those for  $\mathtt{B}$ content versus $\mathtt{D}$ content. For the gray shaded area, $n^*$ is negative which means that the AI's response is all $\mathtt{D}$ content (i.e. bad) from the outset.}   
    \label{fig:3}
\end{figure}

Equation 1 also shows how to prevent output tipping by, for example, increasing $n^*$ beyond the AI's allowed output size in response to a prompt. This can be achieved as shown in Fig. 3(b) for prompt $\mathtt{A}$, by the AI's builder making  ${\vb*S_{A}}\cdot {\vb*S_{B}}$ large; or by a user choosing prompts so that the exponential in the numerator dominates.

Although this same Attention engine empowers all generative AI such as ChatGPT, each commercial LLM (Large Language Model like ChatGPT) has its own additional proprietary features to help improve its performance, including its own proprietary interconnectivity between multiple layers of Attention heads. Nonetheless, in all these cases the $f$ input spins in each iteration pass progressively from the initial to final layers (i.e. $L=1$ to $L_{\rm LLM}$) and the effect of each layer 
is somewhat similar to the single Attention head because the intra-layer Attention heads operate independently in parallel.
The  Attention mathematics means that the outgoing versions of ${\vb*S_{A,B,C,D,\dots}}$ from layer $L$ have shifted alignments and magnitudes compared to the values they had going into layer $L$. Each spin's ingoing and outgoing values for layer $L$ then get added together using some proprietary proportion (called the learning rate in the residual connection) before passing on to layer $L+1$ and the whole process repeats. A layer normalization is also applied to make sure the amplitudes don't trivially scale with $L$. When the final layer $L_{\rm LLM}$ is reached, next token selection occurs as described previously. 
This means that the overall impact on a string of input spins passing through the multilayer structure is that some subsets of spins will become less like the output tipping case (e.g. Fig. 2(a)) by the time they reach $L_{\rm LLM}$ while others will become more like it. This suggests that the likelihood of output tipping occurring in  any multilayer generative AI such as ChatGPT will be very crudely similar to the single Attention head case analyzed above. 

We have also identified an amplification mechanism for this output tipping that will operate exclusively in cutting-edge ChatGPT etc. because of their very large numbers of layers and prompt tokens, and their very large size vocabulary. 
The underlying cause is illustrated in Fig. 4(a) which presents a numerical calculation of the trajectories of the initial spins (tokens) in Fig. 2(a) as they pass through a 10-layer Attention system that includes the realistic LLM features of residual connections, non-identity learned matrices $W$ from the training stage, and layer normalization (see SM for code). The separations of the tokens (spin vectors) change as they pass through successive layers (Fig. 4(a)) with pairs $\mathtt{A}-\mathtt{B}$, $\mathtt{A}-\mathtt{D}$, $\mathtt{B}-\mathtt{D}$  coming closer together (fusion) but $\mathtt{A}-\mathtt{C}$, $\mathtt{B}-\mathtt{C}$, $\mathtt{C}-\mathtt{D}$ moving further apart (fission). 
This means that even if the bad content $\mathtt{D}$ starts off far from the prompt  $\mathtt{A}$ and good content $\mathtt{B}$ in the $d\gg 1$ dimensional embedding space, they end up quite close in the final layer -- perhaps in the same few-dimensional subspace as in Fig. 2(a). Hence Fig. 2(a) may indeed represent a realistic scenario for the final layer of a commercial LLM, as opposed to being a toy model.  

To show how this fusion can then act as a macroscale amplifier of the output tipping, we start by assigning a link between pairs of tokens if their separation is smaller than some threshold. 
Hence the passage of $N\gg 1$ spins (tokens) through the LLM (i.e. increasing $L$) involves links forming and breaking between pairs and hence clusters forming and breaking up (fusion and fission). To account for the large vocabulary, 
we allow the $N$ tokens to have any number of major differences --  i.e. the vocabulary comprises $D$ different `species' of token (e.g. completely different topics or languages) where $D$ can be arbitrarily large -- as well as more minor differences within species. 
The effect of a realistic (i.e. large, language-rich) prompt passing through a realistic LLM is therefore broadly equivalent to the fusion-fission dynamics of a population of $N\gg 1$ heterogeneous objects (different spins) in which successive layers $L$ play the role of successive timesteps.  
Extending the result of Ref. \cite{multispecies}, this clustering follows an inviscid Burgers' equation (see End Matter)
where $N_u$ is the species-$u$ subpopulation: and it can have a shockwave solution that corresponds to the formation of a giant cluster (i.e. a macroscopic giant connected component) with size
\begin{equation}
    G(L) = 1 - \frac{1}{N} \sum_{s=1}^D N_s e^{-2\sum_{r=1}^D \sum_{L'=1}^L F_{sr}(L') G_r /N}\ \ .
\end{equation}
\noindent The implicit summation of different species' fusion and fission contributions predicts the possibility of kinks and dips in this giant cluster's size (Fig. 4(b)) and hence an implicit ongoing competition between macroscale amplification and non-amplification. If fusion dominates, giant (i.e. macroscale) multi-species clusters can form and  act as macroscopic `super-tokens' that bring together good and bad content in a low-dimensional embedding subspace akin to Fig. 2(a) hence making output tipping in the last layer more likely. 
This is exactly what is seen in a multilayer LLM simulation (Fig. 4(c)). 

Our cluster theory equation also predicts  a necessary condition for a giant cluster to form and hence for amplification to be likely: the onset layer $L_c\approx N/2\bar F$ must be less than $L_{\rm LLM}$ where $\bar F$ is an average heterogeneity factor over all token pairs and all layers. This means that output tipping amplification is far more likely to occur in the very large commercial LLMs where the number of layers ($L_{\rm LLM}$) is likely to be much larger than $L_c\approx N/2\bar F$. More generally, fission will compete with fusion, as shown in the numerical results in Fig. 4(c) which also indicate that this competition and hence the amount of amplification will depend on  the size of the embedding dimension compared to the number of tokens involved. We find it particularly intriguing that the Fig. 4(b)(c) shapes are similar to existing curves reported for grokking during AI learning \cite{multispecies,cluster,percolation}.

\begin{figure}[ht]
    \centering
\includegraphics[width=1.0\linewidth]{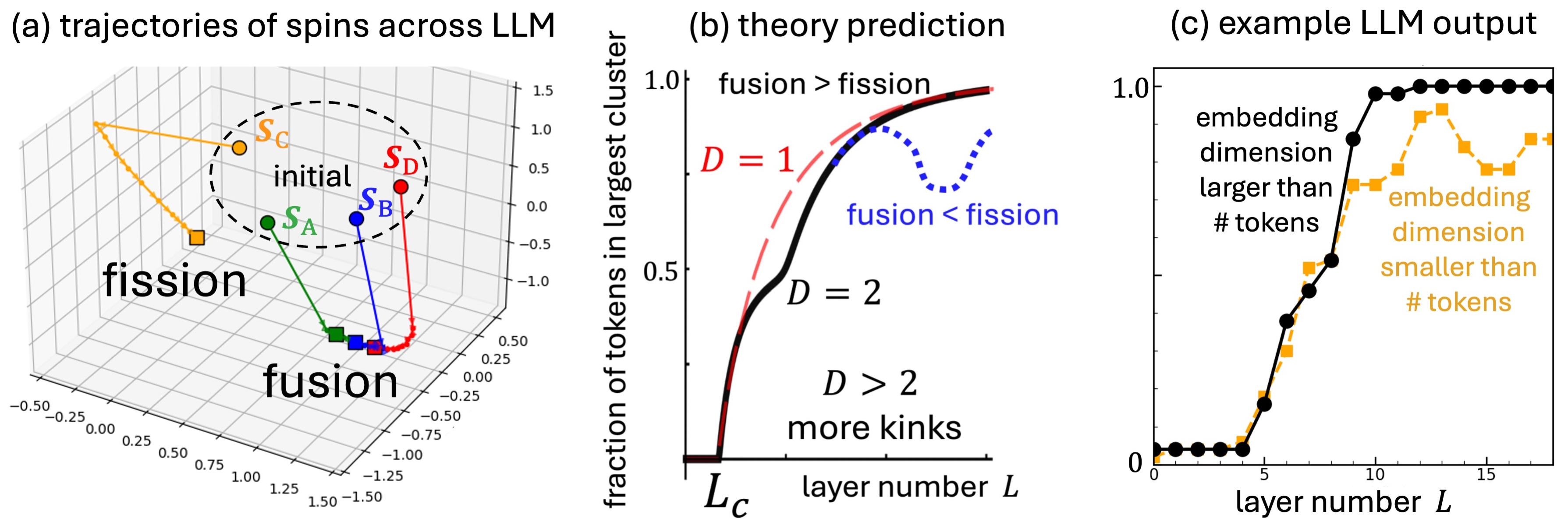}
    \caption{Effect of multiple Attention-head layers, as in commercial LLMs. (a) Trajectories of initial spin vectors (tokens) used in Fig. 2(a). As they pass though to layer $L=10$, $\mathtt{A,B,D}$ become closer (i.e. effective fusion) but $\mathtt{C}$ becomes less close (i.e. effective fission). (b) and (c) compare the theoretical prediction from Eq. 2 to example output from an LLM numerical simulation that incorporates the token fusion and fission shown in panel (a). Curves show the size $G(L)$ of the largest cluster (giant connected component). See SM for the code.}
    \label{fig:4}
\end{figure}

Taken together, our results offer a unified physics understanding and quantitative theory of ChatGPT-like generative AI including potentially harmful hallucinations: from its microscale Attention to its macroscale multilayer complexity. But going further, our results also suggest concrete AI design improvements. For example, the SM shows how 2 new design strategies that follow on directly from our multispin results, do improve performance when applied on a simple GPT-2 model benchmark:
(1) Gap cooling: following Eq.\ 1, increase the gap between the top 2 pairs of interactions when they become too close (i.e.\ just before tipping).  
(2) Temperature annealing: control the temperature dial $T'$ to balance between the risks of output tipping and excessive output randomness. The SM contains full details and code.

\bibliography{references}

\onecolumngrid

\vskip1in 
\begin{center}
    {\large \textbf{End Matter}}
\end{center}
\twocolumngrid

Equation 1 is derived in detail in the SM in step-by-step tutorial style. Equation 2 is a generalization of the result in Ref. \cite{multispecies} to which we refer for full details. If fission becomes extremely infrequent, giant multi-species clusters can suddenly emerge at layer $L_c$.  Their mathematical `shock' shape is due to smaller but substantial clusters fusing together in quick succession. The different token species form the components of a vector generating function $\mathcal{E}$ that obeys 
a generalized $D$-species form of the inviscid Burgers' equation in which $L$ plays the role of time, i.e.
\begin{equation} \label{eq:burgers_Dd}
\partial_L{\mathcal{E}}_s(\vec{y},L) + \frac{2}{N^2}F_{uv}(L)[\mathcal{E}_u - N_u]\partial_v\mathcal{E}_s = 0
\end{equation}
in component form where $N_u$ is the species-$u$ token population and $F_{uv}(L)$ is the average interaction (i.e. average dot-product and hence similarity) between any two tokens from species $u$ and $v$, averaged over all tokens within each species, at layer $L$. The standard approach of characteristics yields the solution 
\begin{equation}
    \mathcal{E}_s = N_s e^{-2 \sum_r \int_0^L \dd L' F_{sr}(L') [N_r-\mathcal{E}_r] /N^2}
\end{equation}
for continuous variable $L$. For discrete and finite number of layers $L$, the growth curve can then be expressed analytically: 
{\small\begin{align}
    G(L) &= \frac{N-\sum_{s=1}^D \mathcal{E}_s(0,L)}{N} \nonumber \\
    &= 1 - \frac{1}{N} \sum_{s=1}^D N_se^{-2 \sum_{r=1}^D \sum_{L'=1}^L F_{sr}(L') G_r /N}
\end{align}}

\noindent where $G_r(L) = (N_r-\mathcal{E}_r(0,L))/N$ is the species composition of the giant cluster, as in Eq. 2.

\end{document}